\documentclass[10pt,conference]{IEEEtran}

\usepackage{xspace}
\usepackage{graphicx}
\usepackage{caption}
\usepackage{algorithm}
\usepackage{algpseudocode}  
\usepackage{amsmath,amssymb}
\usepackage{enumitem}
\usepackage{subcaption}
\usepackage{multirow}
\usepackage{xcolor}
\usepackage{textcomp}
\usepackage[hyphens]{url}
\usepackage{fancyhdr}
\usepackage{hyperref}

\newcommand{\etal}{{\em et al. }}
\newcommand{\BfPara}[1]{{\noindent {\bf #1}}}

\expandafter\def\expandafter\normalsize\expandafter{%
    \normalsize%
    \setlength\abovedisplayskip{5pt}%
    \setlength\belowdisplayskip{5pt}%

}
\newcommand{\Sysname}{DYNAMIX\xspace}
\newcommand{\keqiang}[1]{\textcolor{red}{\textbf{[#1]}}}

\newcommand{\hpcayear}{2026}

\title{DYNAMIX: RL-based Adaptive Batch Size Optimization in Distributed Machine Learning Systems}

\author{
  \IEEEauthorblockN{Yuanjun Dai}
  \IEEEauthorblockA{Case Western Reserve University\\ \texttt{yxd429@case.edu}}
  \and
  \IEEEauthorblockN{Keqiang He}
  \IEEEauthorblockA{Shanghai Jiao Tong University\\ \texttt{kqhe@cs.sjtu.edu.cn}}
  \and
  \IEEEauthorblockN{An Wang}
  \IEEEauthorblockA{Case Western Reserve University\\ \texttt{axw474@case.edu}}
}

\author{
  \IEEEauthorblockN{\itshape Yuanjun Dai}
  \IEEEauthorblockA{Case Western Reserve University\\ \texttt{yxd429@case.edu}}
  \and
  \IEEEauthorblockN{\itshape Keqiang He}
  \IEEEauthorblockA{Shanghai Jiao Tong University\\ \texttt{kqhe@cs.sjtu.edu.cn}}
  \and
  \IEEEauthorblockN{\itshape An Wang}
  \IEEEauthorblockA{Case Western Reserve University\\ \texttt{axw474@case.edu}}
}

\fancypagestyle{camerareadyfirstpage}{%
  \fancyhead{}
  
  \fancyhead[C]{
    \ifdefined\aeopen
    \parbox[][12mm][t]{13.5cm}{\hpcayear{} IEEE International Symposium on High-Performance Computer Architecture (HPCA)}    
    \else
      \ifdefined\aereviewed
      \parbox[][12mm][t]{13.5cm}{\hpcayear{} IEEE International Symposium on High-Performance Computer Architecture (HPCA)}
      \else
      \ifdefined\aereproduced
      \parbox[][12mm][t]{13.5cm}{\hpcayear{} IEEE International Symposium on High-Performance Computer Architecture (HPCA)}
      \else
      \parbox[][0mm][t]{13.5cm}{\hpcayear{} IEEE International Symposium on High-Performance Computer Architecture (HPCA)}
    \fi 
    \fi 
    \fi 
    \ifdefined\aeopen 
      \includegraphics[width=12mm,height=12mm]{ae-badges/open-research-objects.pdf}
    \fi 
    \ifdefined\aereviewed
      \includegraphics[width=12mm,height=12mm]{ae-badges/research-objects-reviewed.pdf}
    \fi 
    \ifdefined\aereproduced
      \includegraphics[width=12mm,height=12mm]{ae-badges/results-reproduced.pdf}
    \fi
  }
  \fancyfoot[C]{}
}
\begin{document}
\maketitle

\ifdefined\hpcacameraready 
  \thispagestyle{camerareadyfirstpage}
  \pagestyle{empty}
\else
  \thispagestyle{plain}
  \pagestyle{plain}
\fi

\newcommand{\hpcaheight}{0mm}
\ifdefined\eaopen
\renewcommand{\hpcaheight}{12mm}
\fi

\begin{abstract}
Existing batch size selection approaches in distributed machine learning rely on static allocation or simplistic heuristics that fail to adapt to heterogeneous, dynamic computing environments.  
We present \Sysname, a reinforcement learning framework that formulates batch size optimization as a sequential decision-making problem using Proximal Policy Optimization (PPO). 
Our approach employs a multi-dimensional state representation encompassing network-level metrics, system-level resource utilization, and training statistical efficiency indicators to enable informed decision-making across diverse computational resources.
Our approach eliminates the need for explicit system modeling while integrating seamlessly with existing distributed training frameworks.
Through evaluations across diverse workloads, hardware configurations, and network conditions, \Sysname achieves up to 6.3\% improvement in the final model accuracy and 46\% reduction in the total training time. Our scalability experiments demonstrate that \Sysname maintains the best performance as cluster size increases to 32 nodes, while policy transfer experiments show that learned policies generalize effectively across related model architectures.
\end{abstract}

\maketitle

\section{Introduction}

Distributed machine learning (DML) has emerged as the predominant paradigm for training increasingly complex models on expansive datasets.
As model architectures grow in parameter count and computational demands, practitioners increasingly rely on distributed training across multiple computational nodes to maintain feasible training timelines. 
Within this paradigm, batch size selection represents a critical hyperparameter that significantly influences both training efficiency and model convergence properties. 
While larger batch sizes generally improve hardware utilization through increased parallelism, they may adversely affect statistical efficiency, potentially degrading convergence rates and generalization performance~\cite{keskar2016large, smith2017don}.

The optimization complexity intensifies substantially in heterogeneous distributed environments, characterized by variance in computational capabilities, network characteristics, and hardware specifications across training nodes.
These heterogeneous configurations arise from several practical considerations: cost optimization through spot instance utilization~\cite{harlap2017proteus}, consolidation of diverse hardware generations within organizational clusters~\cite{hazelwood2018applied}, and workload deployment in multi-tenant infrastructure~\cite{jeon2019analysis}.
Under such conditions, the conventional approach of uniform batch size allocation frequently leads to suboptimal resource utilization, as demonstrated by Jia \etal~\cite{jia2019optimizing}, who observed significant throughput degradation due to synchronization barriers in heterogeneous clusters.

Existing approaches to batch size optimization in distributed environments fall into several distinct categories, each exhibiting particular limitations. 
Static batch size selection methods, as employed by Goyal \etal~\cite{goyal2017accurate}, establish predetermined values through empirical evaluation processes that cannot adapt to dynamic runtime conditions.
Heuristic-based approaches such as You \etal~\cite{you2019large} employ analytical models to scale batch sizes according to system configurations, but typically optimize for singular objectives without considering the interdependencies between hardware efficiency and statistical convergence. 
Recent systems like  BytePS~\cite{jiang2020unified} developed by Jiang \etal and TensorFlow AutoShard~\cite{tfdataex89:online} implement parameter synchronization optimizations, yet primarily focus on communication efficiency rather than comprehensive resource adaptation across heterogeneous nodes.

Semi-dynamic load balancing techniques, proposed by Chen \etal~\cite{chen2020semi}, have demonstrated enhanced performance in non-dedicated environments by adjusting worker loads at iteration boundaries. 
However, these approaches rely on analytical performance models that require explicit system modeling and fail to capture the complex, non-stationary dynamics of distributed training environments.
Similarly, Mirhoseini \etal~\cite{mirhoseini2017device} employed reinforcement learning for device placement optimization but did not address the fundamental batch size optimization problem.
FlexFlow~\cite{jia2019beyond} optimizes parallel execution strategies but maintains fixed batch sizes across workers, limiting adaptability in heterogeneous environments.

In comparative analysis, existing solutions exhibit three critical limitations: 
(1) insufficient adaptability to dynamic operating conditions, with most approaches establishing static configurations or employing simplistic adaptation heuristics; 
(2) inadequate consideration of multi-objective optimization tensions between computational efficiency and statistical convergence; and 
(3) reliance on explicit system modeling that fails to capture the non-stationary nature of distributed training environments, particularly under resource contention and network variability conditions.

To address these fundamental limitations, we introduce \Sysname, a comprehensive framework that reconceptualizes batch size selection as a sequential decision-making problem under uncertainty, employing reinforcement learning to dynamically optimize batch allocations across heterogeneous worker nodes. 
Our approach distinguishes itself through a multi-dimensional state representation that encompasses network-level metrics (throughput, retransmission counts), system-level resource utilization (memory, CPU time ratios), and training statistical efficiency indicators (batch accuracy, gradient statistics). 
By formulating the optimization objective as a reward function that mathematically balances model quality, training efficiency, and system resource constraints,
\Sysname enables holistic optimization across diverse and dynamically evolving computational landscapes.

Our primary contributions include:
\begin{enumerate}[leftmargin=15pt,topsep=5pt]
    \item[1] A mathematical formulation of dynamic batch size optimization that captures the multi-objective nature of the optimization landscape while accounting for both local system characteristics and global training dynamics in distributed environments.
    \item[2] A comprehensive state representation framework that integrates data across network, system, and training domains, enabling robust decision-making across heterogeneous computational resources with varying capabilities and performance characteristics.
    \item[3] A reinforcement learning methodology based on Proximal Policy Optimization (PPO) that efficiently learns batch size adjustment policies through environmental interaction, eliminating requirements for explicit modeling of complex system dynamics while demonstrating consistent empirical convergence across diverse scenarios.
    \item[4] An extensible implementation architecture that integrates with established distributed training frameworks while introducing minimal operational overhead.
\end{enumerate}

Our comprehensive evaluation demonstrates \Sysname's effectiveness across multiple dimensions: We achieve up to 6.3\% improvement in final model accuracy and 46\% reduction in training time compared to static baselines. 
Scalability experiments across 8, 16, and 32 nodes show that \Sysname maintains superior performance as cluster size increases, achieving 92.6\% accuracy versus 81.3\% for static approaches at 32 nodes. 
Policy transfer experiments demonstrate that learned policies generalize effectively within model families without retraining.
Cross-platform validation confirms framework-agnostic capabilities across Ring All-Reduce and BytePS~\cite{jiang2020unified} parameter server architectures with heterogeneous GPU configurations.

\section{Background and Related Work}

\subsection{Distributed machine learning systems}
Distributed machine learning systems have evolved significantly to address the computational demands of training increasingly complex models. 
These systems typically employ data parallelism, model parallelism, or hybrid approaches to distribute workloads across computational nodes~\cite{dean2012large}.
In data-parallel training, which represents the predominant paradigm for contemporary deep learning, each worker maintains a complete model replica while processing distinct subsets of training data~\cite{chilimbi2014project}.
Workers periodically synchronize parameter updates through either centralized parameter servers~\cite{li2014communication} or decentralized all-reduce operations~\cite{sergeev2018horovod}.
The synchronization mechanism significantly influences training dynamics, with Bulk Synchronous Parallel (BSP) representing the most widely adopted approach due to its convergence guarantees~\cite{chen2016revisiting}.
Under BSP, workers synchronize at iteration boundaries through global barriers, ensuring consistent model states across the distributed environment.
While theoretically optimal for convergence, BSP remains susceptible to performance degradation from straggler effects in heterogeneous environments—a fundamental challenge that \Sysname addresses through adaptive batch size optimization.

\subsection{Batch size selection}
Batch size selection presents a complex optimization problem with substantial implications for both computational efficiency and model convergence properties.
From a computational perspective, larger batch sizes enhance hardware utilization through increased parallelism and amortized communication overhead~\cite{goyal2017accurate}.
However, statistical efficiency considerations introduce countervailing constraints, as empirical evidence suggests that excessively large batch sizes may adversely affect convergence dynamics and generalization performance~\cite{keskar2016large,hoffer2017train}.

Keskar \etal~\cite{keskar2016large} demonstrated that large-batch methods tend to converge to sharp minima of the training function, resulting in degraded generalization capabilities.
Masters \etal~\cite{masters2018revisiting} empirically established that small batch sizes of 2-32 samples often provide improved generalization performance and more effective use of computational resources.
These findings have sparked considerable research into techniques that maintain statistical efficiency while scaling batch sizes, including layer-wise adaptive rate scaling~\cite{you2017scaling}, optimization procedure modifications~\cite{lin2018don}, and gradual batch size increase during training~\cite{smith2017don}.

The complexity of batch size selection intensifies in distributed environments where worker nodes exhibit heterogeneous computational capabilities. 
In such settings, uniform batch size allocation frequently leads to suboptimal resource utilization, as demonstrated by Jiang \etal~\cite{jiang2020unified} and Zhang \etal~\cite{zhang2017poseidon}.
The dynamic nature of these environments \textemdash\xspace particularly in multi-tenant systems or when utilizing spot instances \textemdash\xspace further complicates optimization through the introduction of non-stationary computational conditions~\cite{moritz2015sparknet}.
\section{System Overview}
\label{sec:overview}

\subsection{Design requirements}
Distributed machine learning (DML) workloads present unique challenges for resource optimization, particularly in batch size selection, which significantly impacts both training efficiency and model convergence.
Adaptive batch size adjustment during training offers considerable advantages for resource optimization, especially in heterogeneous environments where computing capabilities vary across nodes.
Therefore, developing a training scheduler that determines the optimal batch size for each training participant becomes essential.
To achieve practical utility, such a scheduler should satisfy several key requirements. 
\textit{RQ1}. It must operate with minimal computational overhead on the target training processes to ensure that the optimization mechanisms themselves do not become performance bottlenecks.
\textit{RQ2}. Furthermore, the system should demonstrate scalability by maintaining consistent performance metrics across different deployment scales.
\textit{RQ3}. Finally, robustness must be ensured through resilient operation under fluctuating workloads and network conditions without compromising effectiveness. 
These requirements collectively establish the foundation for a system that can be effectively deployed across diverse training infrastructures and computational environments.
To address these challenges, we propose to adopt an RL learning approach for the design of the training scheduler by leveraging RL's capability to dynamically optimize decision-making processes through environmental interaction and policy refinement.

\subsection{Key insights}
Our proposed approach is founded on several insights regarding the nature of the DML workloads and the limitations of conventional optimization strategies.

\BfPara{Dynamic Optimization} Batch size optimization presents a sequential decision-making problem with significant environment uncertainty.
Training dynamics evolve continuously based on model architectures, dataset characteristics, convergence stage, and resource availability.
Such a dynamic environment lacks closed-form solutions and exhibits non-stationary properties that make heuristic approaches insufficient.
RL offers a framework for optimization under these conditions because it adaptively refines its policy through environmental interactions without explicit mathematical modeling of complex system dynamics.

\BfPara{Multi-dimensional State and Objective Functions} The optimization landscape for batch size selection encompasses multiple competing objectives\textemdash training convergence rate, generalization performance, hardware utilization efficiency, and communication overhead.
Traditional approaches often prioritize a single dimension or rely on simplistic trade-off functions.
RL provides a natural framework for multi-objective optimization through reward functions that can encode complex relationships between these dimensions.
This approach enables holistic optimization that simultaneously considers system-level performance and statistical learning indicators.

\BfPara{Phase-Aware Adaptation}
Distributed training exhibits distinct phases—initial exploration, primary convergence, and final refinement—each with unique optimization requirements. 
Our empirical observations reveal that optimal batch size configurations vary substantially across these phases, requiring dynamic adjustment capabilities that static approaches fundamentally lack. 
By integrating training progress indicators into the state representation, our RL framework develops time-varying policies that automatically transition between exploration-focused and exploitation-focused batch size configurations.

\subsection{System architecture}
\begin{figure}[h]
    \centering
    \includegraphics[scale=0.25]{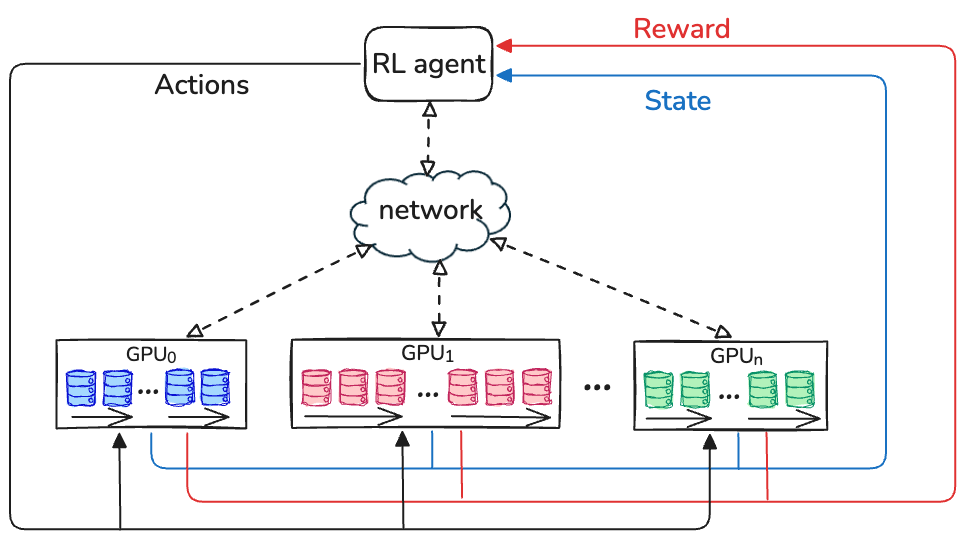}
    \caption{System Architecture}
    \label{fig:arch}
\end{figure}

Figure~\ref{fig:arch} illustrates the overall architecture of our system.
We implement a seamless architecture that integrates RL capabilities with distributed training infrastructure to enable dynamic batch size optimization.
The key component in our system is an RL component that contains a Proximal Policy Optimization~\cite{schulman2017proximal} (PPO)-based learning component that functions as the decision-making engine for batch size adjustment.
It receives comprehensive state information and reward signals from worker nodes.
The agent determines optimal batch size adjustments based on local and global state information.
The intermediate network layer facilitates the communication between the RL agent and distributed workers, handling state information transmission and action distribution.
Note that our framework maintains compatibility with any training platform that implements Bulk Synchronous Parallel (BSP) synchronization, regardless of the specific communication paradigm employed.
This design choice ensures broad applicability across both centralized parameter server architectures (e.g., Parameter Server~\cite{li2014communication}) and decentralized approaches (e.g., Ring All-Reduce), providing platform-agnostic optimization capabilities.

The bottom layer consists of distributed GPU worker nodes with potentially heterogeneous computation capabilities that execute the ML training.
These worker nodes collect and transmit local state information, including system metrics, network conditions, and training statistics, receive batch size adjustment actions from their corresponding RL agents, and implement the adjusted batch sizes for subsequent training iterations.
A key insight incorporated into our design is the temporal aggregation of data across multiple training iterations.
Rather than making decisions based on individual iteration metrics, which often exhibit high variance and limited statistical significance, our system aggregates performance metrics across $k$ consecutive iterations.
This approach yields more reliable indicators of the relationship between batch size and workload states, enabling statistically robust decision-making that captures meaningful performance patterns and filters transient fluctuations.

The system operates in a cyclic manner: after $k$ iterations of training with a given batch size, each worker collects comprehensive state information and transmits it to the RL agent.
The agent then determines the optimal batch size adjustment based on this information, and the worker implements this adjustment for the next $k$ iterations.
Although the RL agent processes local state information from individual worker nodes, it inherently leverages globally shared state information maintained through the BSP synchronization mechanism, such as model generalization capability metrics that remain consistent across the distributed environment.
Such a design strikes a balance between system complexity and optimization efficacy. 
By limiting inter-node communication to synchronization points and utilizing independent RL agents that share implicit global state information, the system maintains scalability while preserving coordination capabilities. 

In the following sections, we provide detailed descriptions of the problem formulation, system components, and implementation details.

\section{Problem Formulation}
\label{sec:formulate}

In this section, we provide a precise mathematical formulation of the problem domain, detailing the state space, action mechanisms, reward structures, and optimization objectives that collectively enable adaptive batch size adjustment.

\subsection{Reinforcement learning framework}
Our system employs a centralized RL paradigm with a single PPO-based agent that coordinates batch size adjustments across all worker nodes.
The system consists of $N$ worker nodes participating in distributed training, each generating local state information.
At each decision point $t$, the centralized RL agent collects both local states $s_t^i$ from each worker node $i$ and the global state $s_t^{global}$ shared across all participants through the BSP-like synchronization mechanism.
The policy for this centralized agent is defined as $\pi_\theta(a_t^i|s_t^i, s_t^{global})$, where $a_t^i$ represents the batch size adjustment for worker node $i$ at time $t$, and $\theta$ denotes the shared policy parameters.
This formulation enables the centralized agent to generate node-specific actions while maintaining parameter consistency across all decisions, effectively balancing individualized optimization with coordinated policy learning.

The agent employs PPO to maximize the expected cumulative reward for each worker node. 
The clipped surrogate objective is expressed as:
\begin{align}
    L^{\text{CLIP}}_i(\theta) = \mathbb{E}_t\Bigl[
\min\Bigl\{ r_t^i \hat{A}_t^i,\, \operatorname{clip}(r_t^i,\,1-\epsilon,\,1+\epsilon)\hat{A}_t^i \Bigr\}
\Bigr]
\label{eq:obj}
\end{align}
, with the probability ratio defined as:
\begin{align*}
    r_t^i = \frac{\pi_{\theta}(a_t^i \mid s_t^i, s_t^{\text{global}})}
             {\pi_{\theta_{\text{old}}}(a_t^i \mid s_t^i, s_t^{\text{global}})}.
\end{align*}
Here $\hat{A}_t^i$ represents the advantage estimate for node $i$ and $\epsilon$ is a hyperparameter controlling the clipping range.
The overall optimization objective is:
\begin{align*}
    J(\theta) = \sum_{i=1}^{N} L^{\text{CLIP}}_i(\theta)
\end{align*}
This centralized approach allows the agent to learn a unified policy applicable to all nodes while still producing individualized batch size adjustments based on node-specific conditions.
In our design, since the action space is very limited and all the components of the reward function are normalized within a stable
range, each episode of training is executed under a relatively consistent environment.
Thus, the reward signal does not fluctuate drastically in this case, allowing us to simplify the PPO algorithm by directly using the cumulative reward for policy updates without relying on the clipping mechanism or explicit advantage estimation. 
While this modification reduces computational overhead, the empirical results demonstrate effective policy learning in our specific application domain.

\subsection{State representation}
For each worker node $i$, the RL agent utilizes a comprehensive state representation that captures multi-dimensional aspects of the training environment.
Each state vector $s_t^i$ is constructed from metrics collected every $k$ iterations and encompasses three distinct categories of information.

\BfPara{Network-level Metrics} 
Network conditions significantly influence distributed training efficiency through their impact on parameter synchronization.
We capture such dynamics by collecting average throughput $Tp_t^i$ and total number of packet retransmissions $Rtx_t^i$ over $k$ iterations.
These metrics provide empirical indicators of network congestion and communication efficiency, enabling adaptation to varying network conditions.
Such metrics have not been sufficiently considered in existing solutions~\cite{chen2020semi}.
However, the importance of these metrics lies in their capability to reveal optimization opportunities obscured by computation-centric performance models.
For example, batch size directly influences synchronization frequency—larger batch sizes lead to fewer synchronization events, thus network throughput and retransmission rates can indicate when to increase batch sizes during congestion periods to reduce communication overhead. 
In heterogeneous networking environments where bandwidth asymmetry, latency variations, and cross-workload traffic introduce stochastic performance characteristics, these network-aware metrics become increasingly critical as model parameter counts continue to scale exponentially.

\BfPara{System-level Metrics} 
Since computational resource utilization may directly affect batch processing capacity, we leverage the CPU and memory utilization to capture computation capacity.
Specifically, we calculate the time ratio between the total CPU time and the wall-clock time over $k$ iterations.
This helps us measure computational efficiency across multi-core systems, where ratios exceeding 1 indicate effective parallel execution.

\BfPara{Training Statistical Efficiency Metrics} 
Model learning dynamics provide critical feedback for batch size optimization.
Therefore, we collect the average $\bar{A}_t^i$ and the standard deviations of batch accuracy $\sigma_{batch,t}^i$, accuracy gain $\Delta A$, and average iteration time $T_\text{iter}$, as part of the state representation.
Among all, $\Delta A$ is computed by first standardizing the batch accuracies (using z-score normalization) and then applying a sliding window-based average.
Then, the difference between the averages of the first and last window is considered accuracy gain.
In practice, adaptive optimizers, such as Adam~\cite{kingma2014adam} or LAMB~\cite{you2019large}, are often used to adjust learning rates to different parameters automatically based on the statistics of gradients. 
Such mechanisms inherently rely on the normalization of
gradient updates, which is critical for ensuring convergence stability
and rapid adaptation to varying data distributions.
To capture the internal dynamics brought by the optimizers, we augment the state representation with two additional metrics, $\sigma_\text{norm}$ and $\sigma_\text{norm}^2$.
They capture the normalized standard deviation of gradients and the normalized variance of gradients, respectively. 
Incorporating these metrics allows the RL agent to gain insight into the scale and stability of the parameter updates induced by the adaptive optimizer.
This comprehensive representation allows the RL framework to evaluate the complex tradeoff between statistical efficiency (quality of updates) and hardware efficiency (throughput), thereby optimizing batch sizes for both convergence rate and computational throughput simultaneously.

The complete state vector for each node is constructed as the concatenations of the abovementioned metrics.
The global state $s_t^{global}$ encompasses shared training metrics such as global loss trajectory, validation accuracy, and model convergence indicators that remain consistent across all nodes due to the BSP synchronization mechanism.

\subsection{Action space and batch size adjustment}
The centralized agent employs a discrete action space that constrains batch size adjustments to manageable increments.
Specifically, $\mathcal{A}={-100, -25, 0, +25, +100}$.

This design choice is derived through systematic sensitivity analysis across multiple model-optimizer combinations. 
Our empirical analysis revealed that continuous action spaces introduce significant training instability, as early-stage policy exploration generates random, large adjustments that lead to gradient variance oscillations, disrupting SGD convergence trajectories. 
This effect is particularly pronounced for adaptive optimizers like ADAM and LAMB, which rely on gradient moment estimates that are sensitive to abrupt batch size changes.
The increment $\pm 100$ enables rapid early-stage adaptation, while $\pm 25$ provides fine-grained mid-training adjustments. 
This granularity balances exploration capability with gradient statistic preservation.
For each worker node $i$, upon receiving an action $a_t^i$, the batch size is updated to $\text{BatchSize}_{t+1}^i = \text{BatchSize}_t^i + a_t^i$.
The updated batch size is further constrained within a predefined range [32, 1024] to ensure computational feasibility and statistical validity.
This constrained discrete action space fluctuations in batch sizes, maintains hardware compatibility by avoiding memory overflows, and ensures statistical representativeness by preventing excessively small or large batch sizes.

\subsection{Reward function design}
The formulation of an effective reward function presents a fundamental challenge in RL-based optimization frameworks.
Our reward function design is motivated by the imperative to balance multiple competing objectives within the distributed learning system while providing empirically stable and informative learning signals to guide batch size optimization.
The reward structure must simultaneously incentivize improvements in model generalization capability, training convergence rate, computational efficiency, and system resource utilization.
Therefore, we consider the following metrics in the design of our reward function.

\BfPara{Generalization Capability} Validation performance is often used to evaluate a model's generalization capability.
However, a complete evaluation using validation datasets can be too expensive to perform frequently during the distributed training of a model.
As a result, we use the mean batch accuracy $\bar{A}_t^i$ over the most recent $k$ iterations as the proxy metric for the target model's current generalization capability.
Furthermore, we also augment it with an amplified accuracy gain $\alpha \cdot max\{0, \Delta A_t\}$ that specifically rewards positive learning trajectory improvements while remaining neutral to temporary performance fluctuations.

\BfPara{Training Efficiency} The iteration time penalty $-\beta \cdot T_\text{iter}$ introduces a computational efficiency dimension, creating direct pressure toward configurations that maximize hardware utilization without compromising statistical efficiency.
This term is particularly significant in heterogeneous computing environments where hardware capabilities vary substantially across nodes.

\BfPara{Batch Size Regularization} To discourage extreme batch sizes,
which may lead to hardware constraints or poor generalization,
we also impose a batch size regularization term $-\delta(log_2(\text{BatchSize}_t)-5)$.
The logarithmic formulation creates a balanced pressure against both excessively large batches and excessively small batches.
The constant value $5$ derives from the minimum batch size of 32.

For adaptive optimizers, we introduce normalization-specific penalty terms $-\eta(\sigma^2_{norm, t} + \sigma_{norm,t})$, that explicitly account for the internal gradient scaling mechanisms inherent in these algorithms. 
These terms help prevent batch size configurations that would destabilize the delicate equilibrium between first and second moment estimates that adaptive optimizers maintain.

Taking all these metrics into consideration, we have the following reward functions for SGD and non-SGD training regimes, respectively:
\begin{align*}
    r_t^{\text{SGD}} = \bar{A}_t + \alpha \cdot \max\{0,\,\Delta A_t\} \nonumber - \beta\,T_{\text{iter},t} - \delta\left(\log_2(\text{BatchSize}_t)-5\right).
\end{align*}
and
\begin{align*}
r_t^{\text{optimizer}} &= \bar{A}_t + \alpha \cdot \max\{0,\,\Delta A_t\} \nonumber - \beta\,T_{\text{iter},t} - \eta\,\Bigl(\sigma^2_{\text{norm},t} + \sigma_{\text{norm},t}\Bigr) \nonumber\\[1mm]
&\quad - \delta\left(\log_2(\text{BatchSize}_t)-5\right).
\end{align*}
To facilitate long-term optimization, we also employ a cumulative discounted reward formulation:
\begin{align*}
    J(\pi) = \mathbb{E}_{\pi}\left[ \sum_{t=0}^{\infty} \gamma^t r_t \right],
\end{align*}
This temporal integration mechanism encourages the RL agent to consider both immediate performance gains and sustained improvements in training efficiency. 
The discount factor $\gamma \in [0,1]$ balances the influence of immediate versus future rewards, providing stability to the learning process while maintaining responsiveness to dynamic environmental conditions.

This comprehensive reward structure enables our framework to navigate the complex, multi-dimensional optimization landscape of distributed training, adapting batch sizes to maximize both model quality and training efficiency under diverse and dynamic computational conditions.

\section{Implementation and Operational Workflow}
\label{sec:impl}
\Sysname implements an end-to-end system for dynamic batch size optimization that can be integrated with existing ML platforms, including Pytorch~\cite{paszke1912pytorch}, Tensorflow~\cite{abadi2016tensorflow} and MXNet~\cite{chen2015mxnet}.
The implementation architecture prioritizes operational efficiency, framework modularity, and integration flexibility with existing distributed training infrastructures.

\BfPara{Key Components} 
At its core, \Sysname consists of three primary modules that work in concert to enable dynamic batch size optimization.
The data collection module gathers multi-dimensional state information from the distributed training environment.
This module employs two complementary collection mechanisms.
For system-level metrics, we leverage the Linux eBPF~\cite{vanbever2019bpf} technology to execute lightweight programs directly within the kernel, avoiding costly user-kernel context switches and enabling direct access to in-kernel data with minimal overhead.
This approach significantly reduces the performance impact of continuous system monitoring compared to traditional system call-based approaches.
Simultaneously, the module collects training efficiency metrics such as batch accuracy and iteration time directly from the training loop. These metrics emerge naturally during the training process and can be captured with negligible overhead.

To facilitate the communication between worker nodes and the centralized RL arbitrator, we leverage the gRPC protocol~\cite{gRPC55:online} to transmit structured data for both state information and batch size adjustment commands.
The communication layer integrates tightly with the data collection mechanism, enabling both components to operate within the same polling loop and further reducing system impact.

The RL Arbitrator serves as the decision-making module of the framework.
It is centralized on a dedicated node to minimize interference with training processes.
Upon receiving aggregated state information from workers, the arbitrator feeds this data into the RL agent, which computes optimal batch size adjustments according to its learned policies. 
These adjustments are then communicated back to worker nodes for implementation in subsequent training iterations.

\BfPara{Deployment Configurations}
The architectural design of \Sysname accommodates multiple deployment configurations to address varying operational requirements and resource constraints. 
For environments with highly utilized worker nodes, a dedicated RL arbitrator node configuration provides the most suitable approach, isolating the computational demands of policy evaluation from the primary training processes. 
In scenarios with asymmetric resource utilization, the framework can be deployed with the RL arbitrator assigned to a low-load node within the cluster, balancing resource utilization while maintaining separation of concerns.

For environments with abundant computational resources, a fully distributed configuration becomes viable, wherein an independent agent resides directly on each worker node. 
This approach eliminates the need for centralized arbitration and associated communication overhead, enabling real-time control and adjustment with minimal latency. 
The synchronization mechanisms inherent in BSP training ensure that batch size adjustments remain coordinated across nodes even in this decentralized configuration.

\BfPara{Operational Workflow}
\Sysname's operational workflow is shown in Algorithm~\ref{alg:workflow}, encompassing several phases.

\begin{algorithm}[ht]
\caption{Independent Learning PPO for Dynamic Batch Size Optimization}
\small
\begin{algorithmic}[1]
\State \textbf{Input:} Worker set \(W\), initial batch size \(\bar{x}\), maximum \(x_{\max}\) and minimum \(x_{\min}\) batch sizes
\State \textbf{Output:} Updated model parameters \(\theta\)
\State

\For{each worker \(i \in W\)}
    \State Initialize data collector and gRPC server
    \State Set batch size: \(x_i \gets \bar{x}\)
    \State Connect to RL arbitrator and signal readiness
\EndFor
\State Wait until all workers are ready
\State

\While{training not converged}
    \For{\(iter = 1 \dots k\)} \Comment{\(k\) iterations per adjustment cycle}
        \For{each worker \(i \in W\) \textbf{in parallel}}
            \State Sample batch \(B_i\) of size \(x_i\)
            \State Compute gradient \(g_i\) and update global model
            \State Collect performance metrics
        \EndFor
    \EndFor
    
    \For{each worker \(i \in W\)}
        \State Aggregate metrics into state vector \(s_i\)
        \State Transmit \(s_i\) to RL arbitrator
    \EndFor
    
    \For{each worker \(i \in W\)}
        \State \(a_i \gets \pi_\theta(s_i, s_{global})\) \Comment{Compute adjustment action}
        \State Update batch size: \(x_i \gets \max\!\bigl(\min(x_i + a_i, x_{\max}), x_{\min}\bigr)\)
    \EndFor
    
    \For{each worker \(i \in W\)}
        \State Compute reward \(r_i\) and advantage \(\hat{A}_i\)
        \State Update policy using Eq~\ref{eq:obj}
    \EndFor
\EndWhile
\State

\State RL arbitrator broadcasts termination signal
\For{each worker \(i \in W\)}
    \State Shut down data collector and gRPC server
\EndFor
\end{algorithmic}
\label{alg:workflow}
\end{algorithm}

\noindent In this workflow, the process begins by initializing each worker node with a data collector, a gRPC server, and an initial batch size, while establishing communication with a centralized RL arbitrator.
Once all nodes signal they are ready, the system enters the main training loop. 
During each cycle, every worker simultaneously samples a mini-batch, computes gradients, and collects performance metrics over a fixed number of iterations leveraging eBPF.
These metrics are then aggregated into state vectors and sent to the RL arbitrator, which computes adjustment actions for each node based on both local and global state information. 
The computed actions are used to update the batch sizes within predefined limits, ensuring stability and efficient resource usage.
Additionally, each worker calculates a reward and advantage estimate to refine the policy using a clipped PPO objective. 
This iterative process continues until convergence, after which the arbitrator issues a termination signal, and the system gracefully shuts down all components.

Our implementation preserves the computational efficiency of the distributed training process while introducing minimal overhead through strategic placement of decision points at iteration boundaries. 
By aligning batch size adjustments with the natural synchronization barriers of BSP training, the framework integrates seamlessly with existing distributed training infrastructures while providing substantial performance benefits through dynamic resource allocation.
\section{Evaluations}
\label{sec:evaluations}
We conduct a comprehensive evaluation of \Sysname to address five critical research questions.
ERQ1: Can \Sysname effectively learn adaptive batch size policies through reinforcement learning? 
ERQ2: How does \Sysname compare against static batch size approaches in terms of accuracy and convergence time? 
ERQ3: Does \Sysname maintain or improve performance as cluster size increases?
ERQ4: Can policies learned on one model architecture generalize to related architectures within the same family?
ERQ5: Is \Sysname robust across different distributed training frameworks and heterogeneous hardware configurations?
To answer these questions, we conduct baseline performance analysis, RL agent training and inference evaluation, scalability experiments across multiple cluster sizes, policy transfer experiments between related model architectures, and cross-platform validation using BytePS~\cite{jiang2020unified} with heterogeneous GPU configurations.

\subsection{Experiment Setup and Methodology}

We conduct a comprehensive evaluation of \Sysname across diverse distributed training scenarios using different benchmark datasets to evaluate its effectiveness in optimizing batch size for heterogeneous environments.

\BfPara{Testbed} 
Our experiments utilize three distinct computing environments to validate \Sysname's performance across different scales and hardware configurations:

\BfPara{\textendash\xspace Primary Testbed (Lambda GPU Cloud)~\cite{LambdaGP49:online}}17 nodes with NVIDIA A100 GPU (24GB memory) and CUDA 12.4. Out of these, 16 function as worker nodes using Ring-based All-Reduce, with one dedicated RL agent node.

\BfPara{\textendash\xspace OSC High-Performance Cluster} 
For scalability experiments, we utilize the Ohio Supercomputing Center (OSC)~\cite{OhioSupercomputerCenter1987} high-performance computing environment with NVIDIA A100-PCIE-40GB GPUs. 
We have three cluster configurations: 8 nodes, 16 nodes, and 32 nodes, with an additional dedicated node hosting the \Sysname RL agent.

\BfPara{\textendash\xspace Fabric Testbed}
To validate framework agnosticism and hardware heterogeneity resilience, we deploy experiments on the FABRIC testbed~\cite{fabric-2019} using 8 heterogeneous GPU worker nodes \textendash\xspace 4 equipped with NVIDIA RTX 3090 GPUs and 4 with NVIDIA T4 GPUs \textendash\xspace plus one CPU-only coordination node.


\BfPara{Workloads and Datasets}
Out of the 17 nodes in our primary testbed, 16 function as worker nodes following Ring-based All-Reduce communication paradigm, with the remaining node serving as the dedicated training scheduler and RL agent.
We evaluate VGG and ResNet family model architectures representing different computational complexity levels, including VGG11, VGG16, VGG19, ResNet34, and ResNet50. 
Our experiments utilize CIFAR-10~\cite{krizhevsky2009learning} and CIFAR-100~\cite{cifar100} datasets, evaluating with both SGD and ADAM optimizers to demonstrate \Sysname's effectiveness across different optimization algorithms.

For distributed training communication, we employ multiple paradigms: PyTorch's DistributedDataParallel (DDP) with Gloo backend for our primary experiments, NCCL backend with Ring All-Reduce for scalability experiments, and BytePS parameter server architecture for framework agnosticism validation. 
Data partitioning is performed using DistributedSampler to ensure balanced distribution across worker nodes.

\BfPara{Metrics} 
To comprehensively assess \Sysname's performance, we employ metrics that capture both computational efficiency and statistical learning effectiveness, including final model accuracy, total convergence time, batch size adaptation, convergence trajectory stability, and policy transferability.

\subsection{Baseline}
\label{sec:baseline}
To establish performance baselines and understand the impact of static batch size selection on distributed training, we first conduct a systematic evaluation of traditional Bulk Synchronous Parallel (BSP) training with fixed batch sizes.
These results serve as the foundation for evaluating the improvements provided by \Sysname's dynamic batch size optimization.

\begin{figure*}[t]
    \centering
    \begin{subfigure}[t]{0.24\linewidth}
        \centering
        \includegraphics[width=\linewidth]{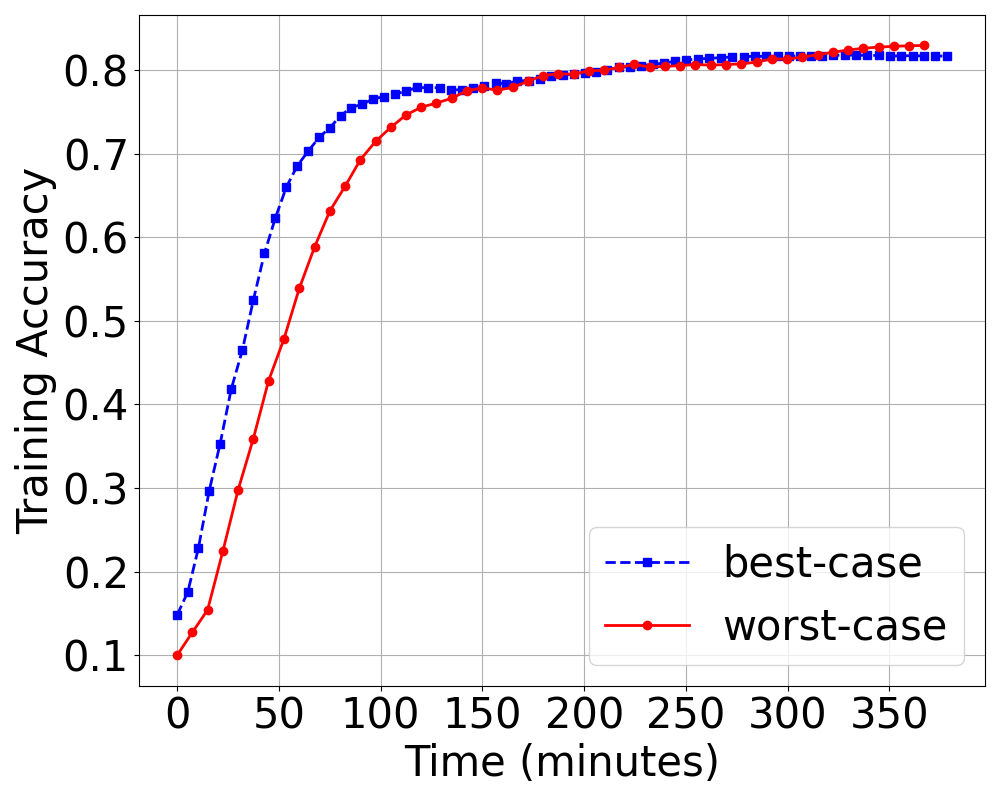} 
        \caption{VGG11 on CIFAR-10 with SGD (Batch Size=32)}
        \label{fig:sgd_32_baseline}
    \end{subfigure}%
    \hfill
    \begin{subfigure}[t]{0.24\linewidth}
        \centering
        \includegraphics[width=\linewidth]{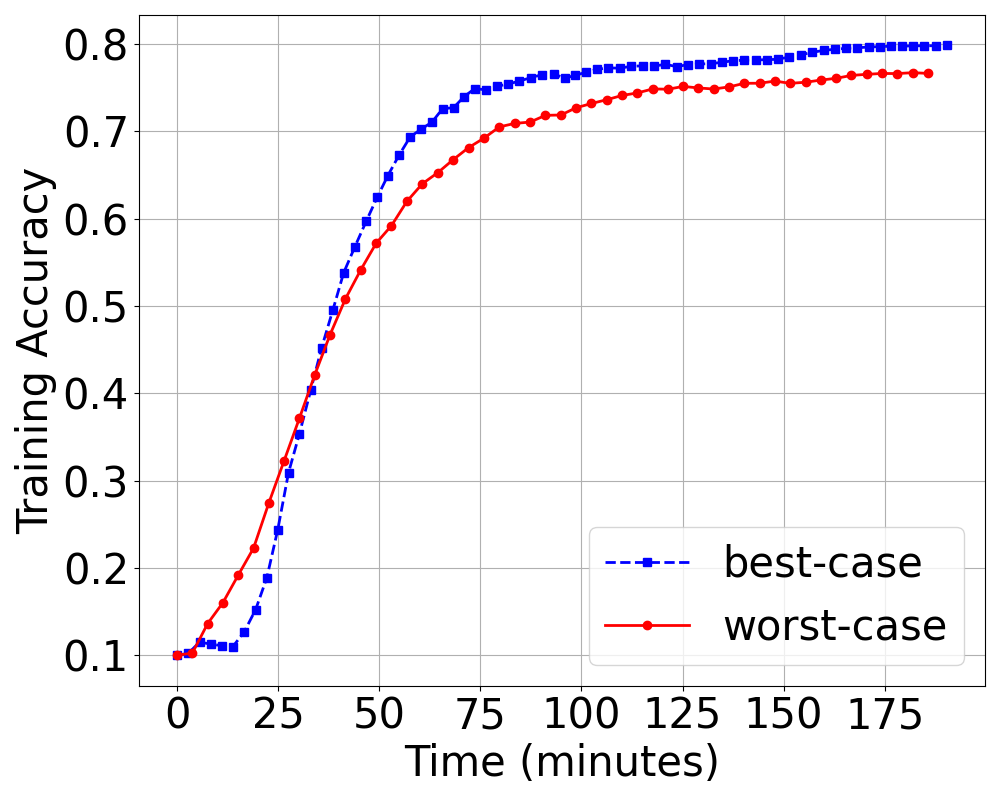} 
        \caption{VGG11 on CIFAR-10 with SGD (Batch Size=64)}
        \label{fig:sgd_64_baseline}
    \end{subfigure}%
    \hfill
    \begin{subfigure}[t]{0.24\linewidth}
        \centering
        \includegraphics[width=\linewidth]{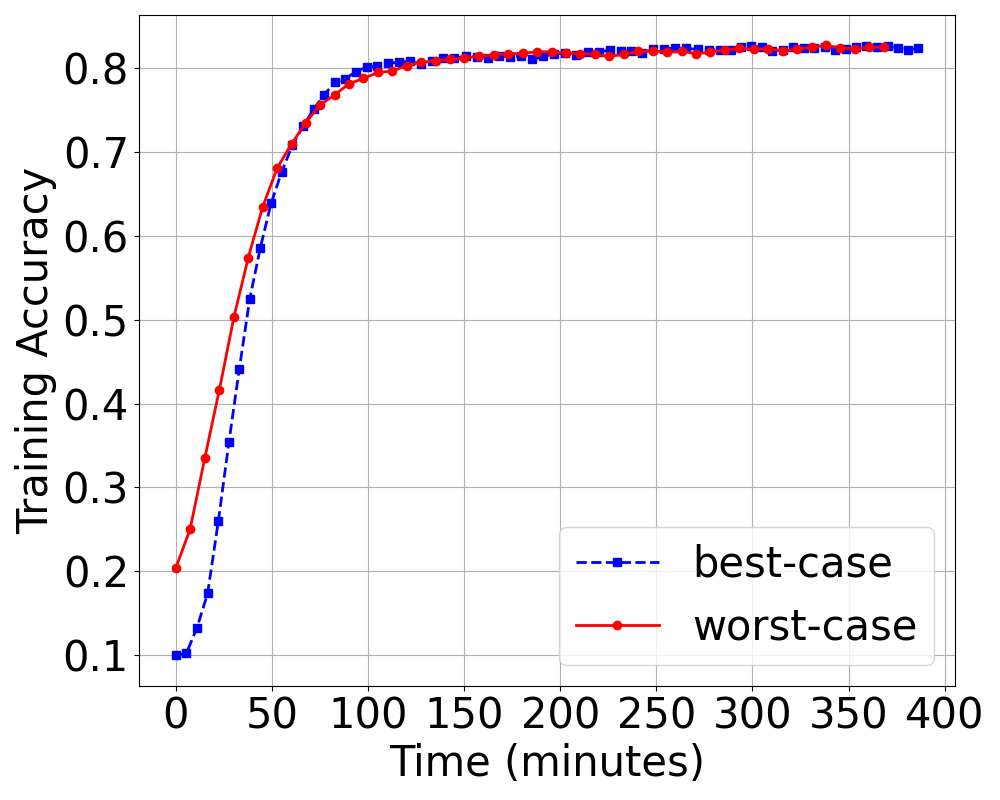} 
        \caption{VGG11 on CIFAR-10 with ADAM (Batch Size=32)}
        \label{fig:adam_32_baseline}
    \end{subfigure}%
    \hfill
    \begin{subfigure}[t]{0.24\linewidth}
        \centering
        \includegraphics[width=\linewidth]{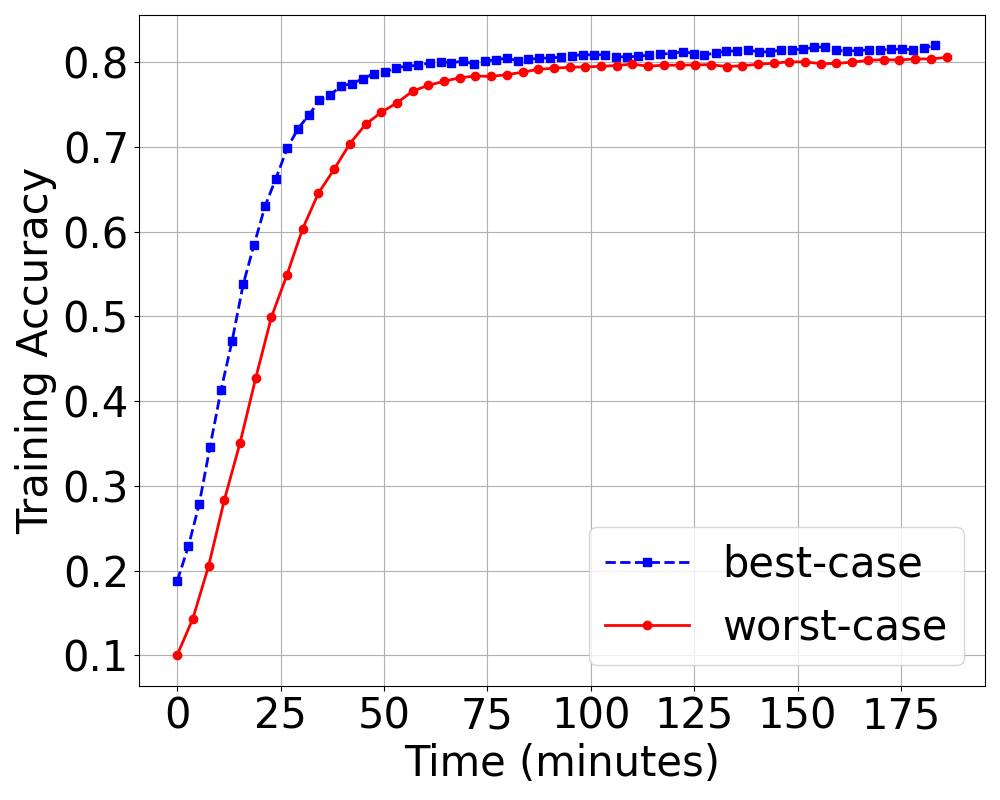} 
        \caption{VGG11 on CIFAR-10 with ADAM (Batch Size=64)}
        \label{fig:adam_64_baseline}
    \end{subfigure}%
    \hfill
    \begin{subfigure}[t]{0.24\linewidth}
        \centering
        \includegraphics[width=\linewidth]{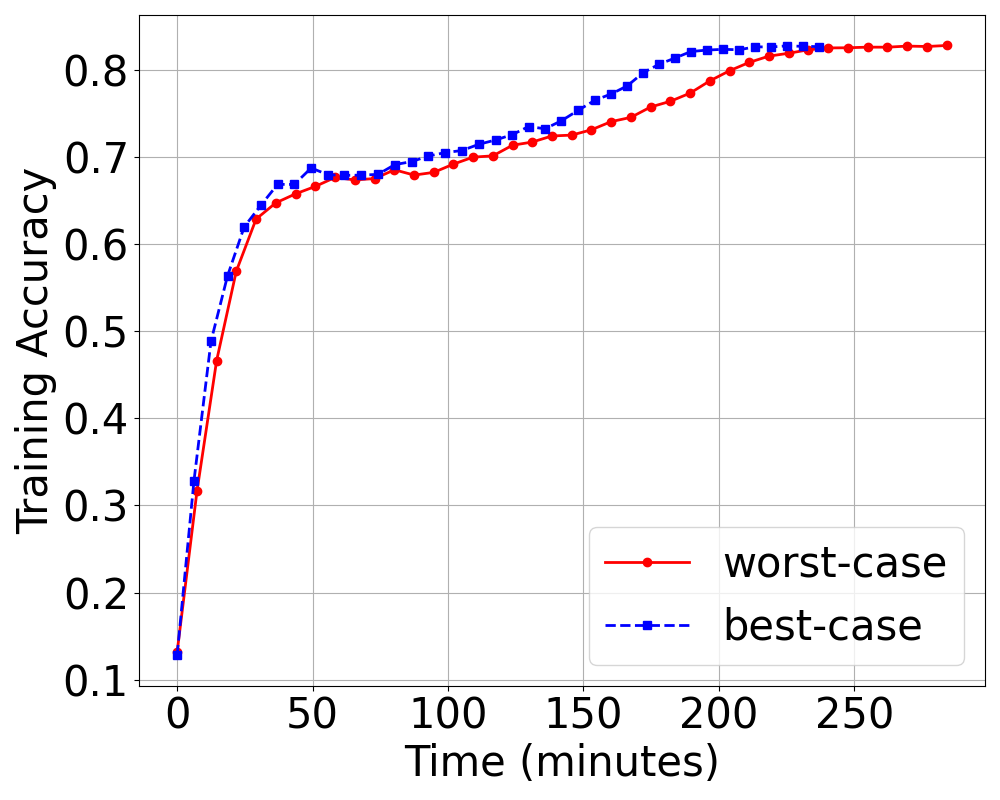} 
        \caption{ResNet34 on CIFAR-100 with SGD (Batch Size=32)}
        \label{fig:resnet_32_baseline}
    \end{subfigure}%
    \hfill
    \begin{subfigure}[t]{0.24\linewidth}
        \centering
        \includegraphics[width=\linewidth]{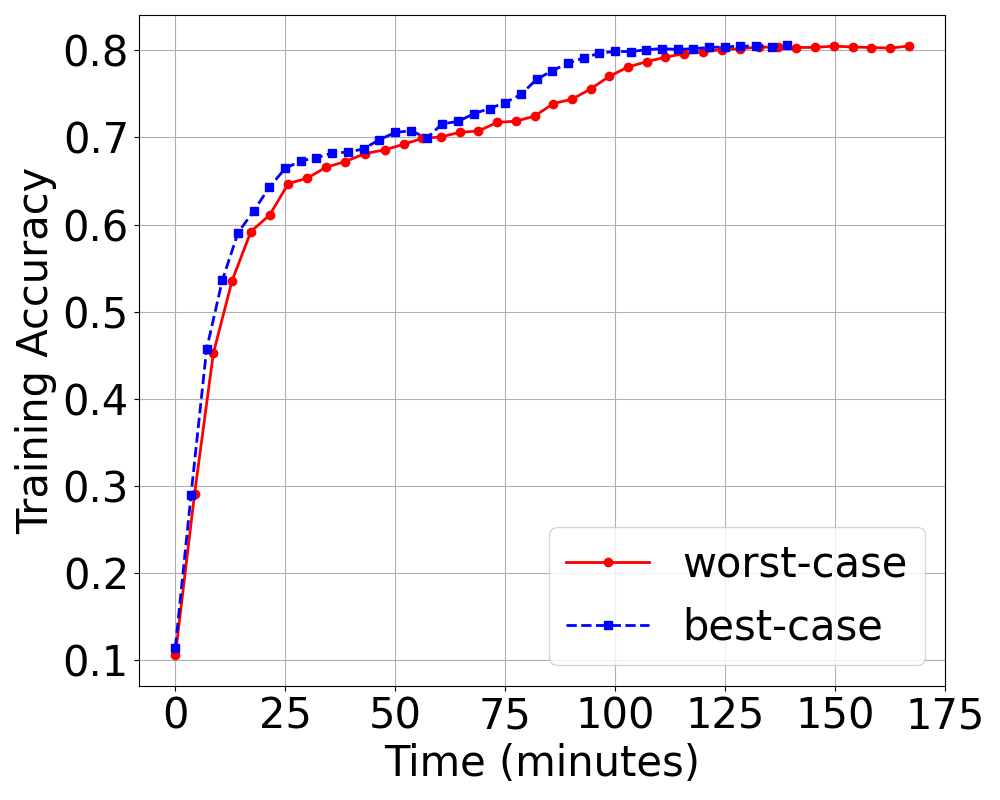} 
        \caption{ResNet34 on CIFAR-100 with SGD (Batch Size=64)}
        \label{fig:resnet_64_baseline}
    \end{subfigure}%
    \hfill
    \begin{subfigure}[t]{0.24\linewidth}
        \centering
        \includegraphics[width=\linewidth]{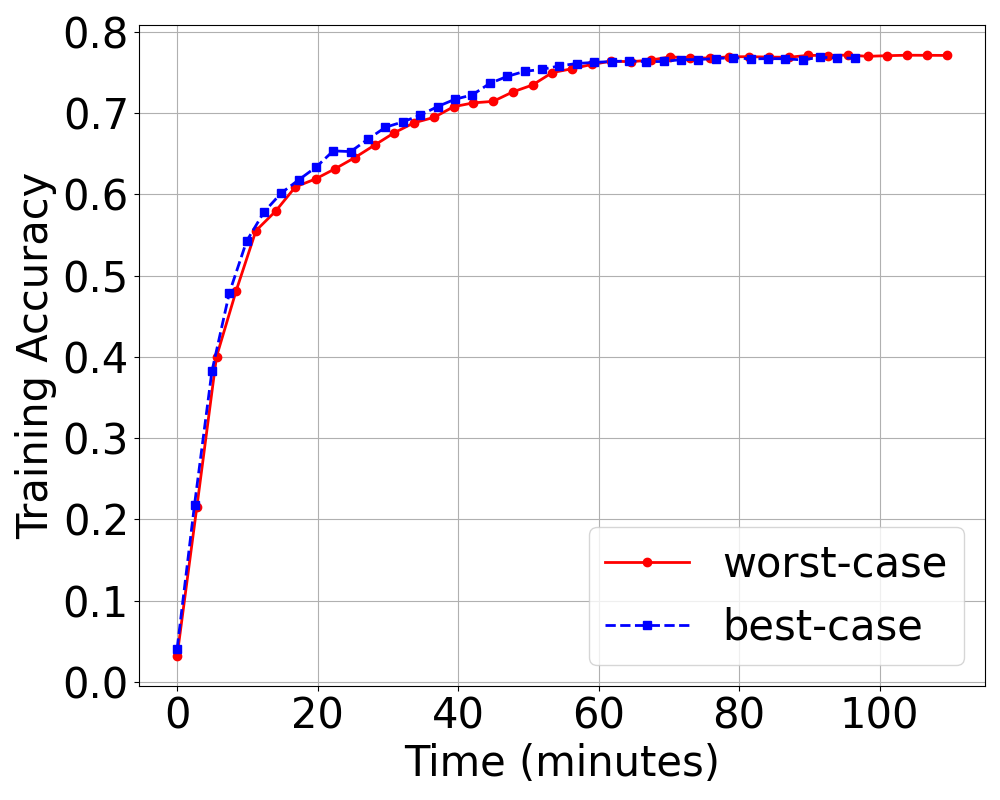} 
        \caption{ResNet34 on CIFAR-100 with SGD (Batch Size=128)}
        \label{fig:resnet_128_baseline}
    \end{subfigure}%
    \hfill
    \begin{subfigure}[t]{0.24\linewidth}
        \centering
        \includegraphics[width=\linewidth]{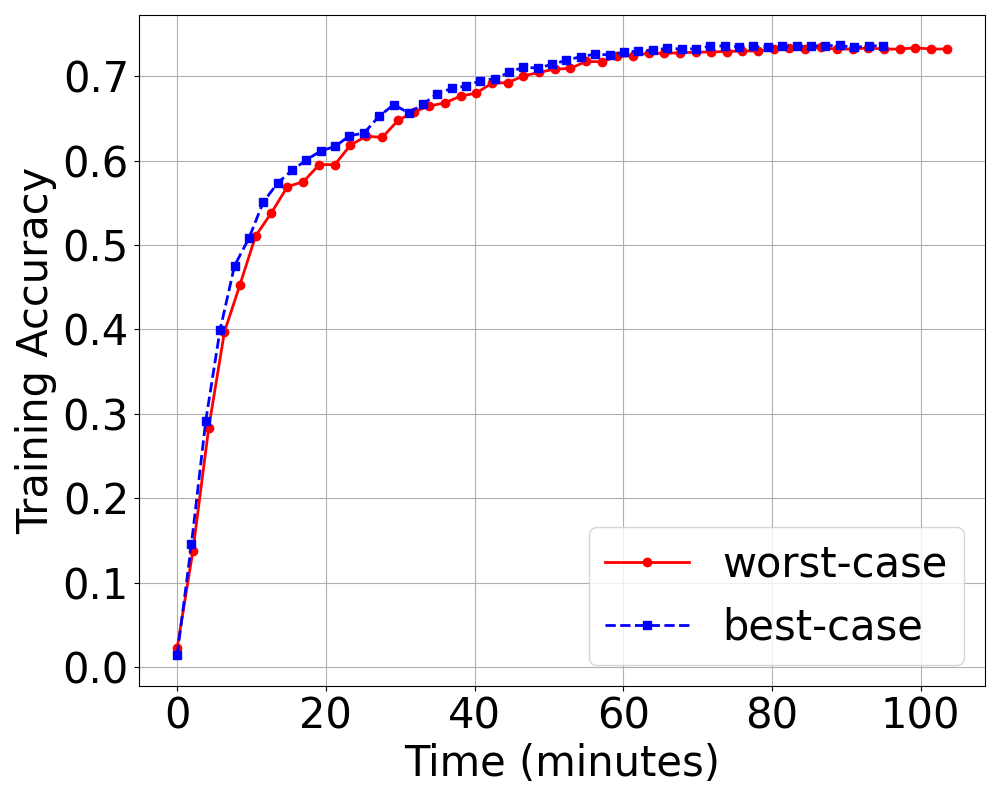} 
        \caption{ResNet34 on CIFAR-100 with SGD (Batch Size=256)}
        \label{fig:resnet_256_baseline}
    \end{subfigure}%
    \caption{Baseline performance with fixed batch sizes}
    \label{fig:baseline}\vspace{-2em}
    \end{figure*}

We systematically evaluate the performance of VGG11 and ResNet34 on the CIFAR-10 and CIFAR-100 datasets, respectively, using a range of static batch sizes (32 - 1024) with both SGD and ADAM optimizers. 
Our primary focus is on identifying configurations that consistently converge to acceptable solutions with good generalization capability.
This evaluation revealed that not all batch sizes yield successful convergence \textemdash larger batch sizes frequently resulted in suboptimal local minima or complete convergence failure, particularly with the Adam optimizer.
Through multiple experimental runs, we find that batch sizes 32 and 64 consistently provide optimal performance for distributed training when using the optimizers.
The results are illustrated in Figure~\ref{fig:baseline}, which depicts convergence trajectories for the successful configurations.

In determining the best- and worst-case scenarios, our primary criterion is the final validation accuracy achieved at convergence. 
For cases where multiple runs demonstrate comparable convergence accuracy (whose discrepancy is within 1\% difference), we use the time needed to reach convergence as the secondary criterion.
Given these criteria, the best-case scenario represents the run that either achieves the highest final accuracy, or when accuracies are effectively equivalent, reaches convergence in the shortest time.
Figures~\ref{fig:sgd_32_baseline} and ~\ref{fig:sgd_64_baseline} show the training trajectory of SGD-based training on VGG11 with static batch sizes of 32 and 64, respectively.

For the model trained with batch size 32 (Figure~\ref{fig:sgd_32_baseline}), we can see that the final training accuracy reaches approximately 0.82 across three runs. The total convergence time spans approximately 350 minutes.
In contrast, the model trained with batch size 64 (Figure~\ref{fig:sgd_64_baseline}), the final training accuracy ranges between 0.76 and 0.79.
However, the total convergence time is nearly half of that in the case of batch size 32.
These comparative results reveal the fundamental trade-off between statistical and computational efficiency in distributed training.
Smaller batch sizes provide better gradient estimates with higher variance, enabling a more thorough exploration of the optimization landscape.
While larger batch sizes allow more efficient hardware utilization through increased parallelism and reduced synchronization overhead.
Similar trends can also be observed in the training trajectories of using the ADAM optimizer as shown in Figures~\ref{fig:adam_32_baseline} and ~\ref{fig:adam_64_baseline}.

To validate that the observed trade-off is not specific to any model architectures or datasets, we extend our baseline experiment to ResNet34 trained on the CIFAR-100 dataset. 
Unlike in VGG11, the training of ResNet34 can reach convergence across a broader range of batch configurations, from 32 to 256. 
Figures~\ref{fig:resnet_32_baseline} to ~\ref{fig:resnet_256_baseline} present the corresponding training trajectories.
By comparing Figure~\ref{fig:resnet_32_baseline} and Figure~\ref{fig:resnet_256_baseline}, we can see that batch size 32 achieves substantially higher model accuracy (0.82) compared to batch size 256 (0.73), despite requiring approximate 2x longer convergence time.
We can also observe that after some inflection point, between batch sizes 128 and 256, additional increases in batch size yield negligible reductions in convergence time while imposing significant penalties on model generalization capability.
The performance variability observed within each batch size configuration further underscores the limitations of fixed batch size approaches.
Furthermore, even with identical hyperparameters, convergence trajectories exhibit significant run-to-run variance, particularly in early training phases. 
This variability stems from the complex interplay of stochastic optimization dynamics, hardware resource fluctuations, and network conditions \textemdash\xspace factors that static batch allocation strategies cannot adapt to during training.
These baseline results establish a clear motivation for dynamic batch size optimization approaches like \Sysname.

\subsection{RL Agent Training Analysis}
\label{sec:rl_training}
To empirically validate \Sysname's effectiveness, we conduct systematic training of the RL agent across multiple experimental configurations. 
This section presents key findings from the training process, highlighting convergence properties and optimization effectiveness.

\begin{figure}[h]
  \centering
  \begin{subfigure}[t]{0.49\columnwidth}
    \centering
    \includegraphics[width=\columnwidth]{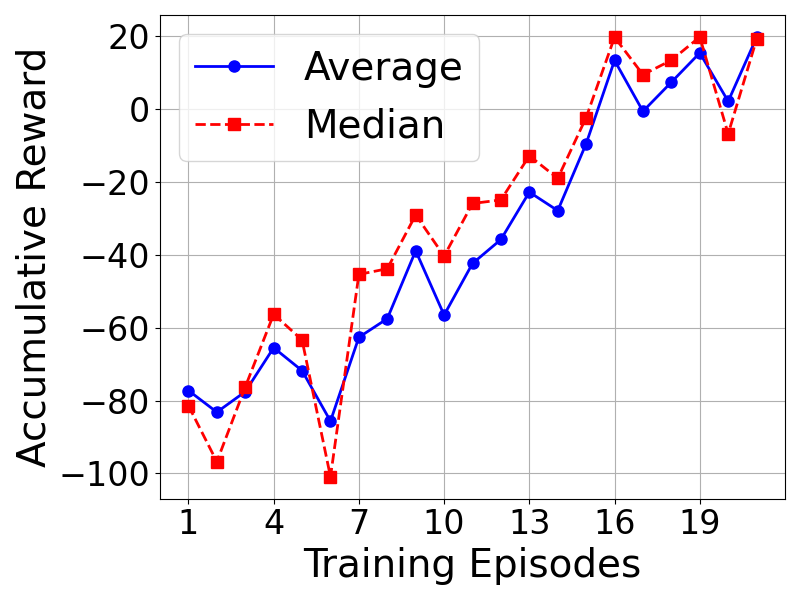} 
    \caption{VGG11 with ADAM Optimizer}
    \label{fig:vgg11_adam_reward}
  \end{subfigure}\hfill
  \begin{subfigure}[t]{0.49\columnwidth}
    \centering
    \includegraphics[width=\columnwidth]{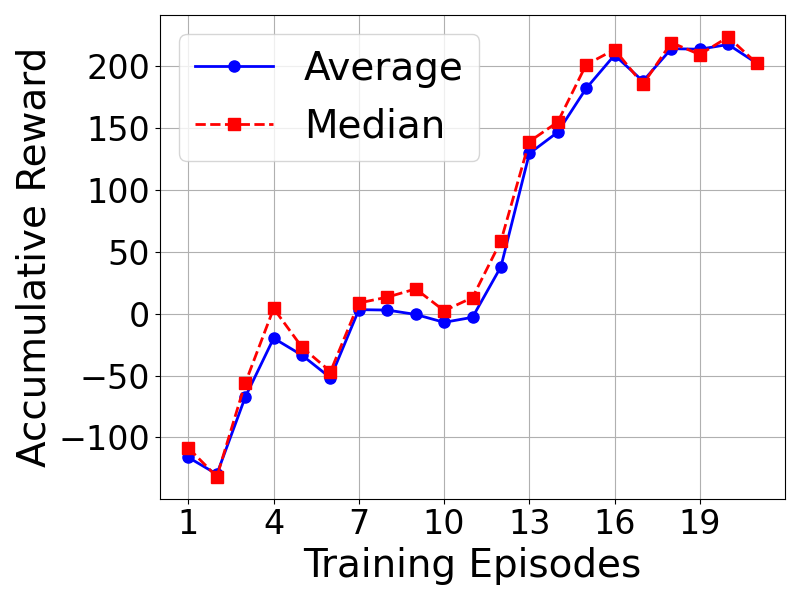} 
    \caption{ResNet with SGD Optimizer}
    \label{fig:resnet_sgd_reward}
  \end{subfigure}
  \caption{Average and median accumulative rewards}
  \label{fig:accu_rewards}
\end{figure}

We train separate RL agents for each distinct configuration.
Each agent goes through 20 training episodes, which is sufficient for policy convergence based on our empirical observations.
During each training episode, the target models are trained with a fixed number of steps, which are derived empirically to guarantee the target model's convergence.
For the VGG11 model, we train 100 steps per episode when using the SGD optimizer, while the ADAM optimizer only requires 70 steps per episode to achieve comparable convergence.
For the ResNet34 model trained with the SGD optimizer, it needs 120 steps per episode to guarantee model convergence.
Note that the number of steps also corresponds to the number of learning steps of the RL model per episode.
At the beginning of each episode, all model weights, optimizer states, and system configurations were reset to initial conditions to ensure unbiased learning.
We demonstrate the average and median accumulative rewards trajectories of the training of VGG11 and ResNet34 over CIFAR-10 and CIFAR-100, respectively, in Figure~\ref{fig:accu_rewards}.
In our experiments, we note that individual worker nodes initially exhibit considerable variance in reward patterns, reflecting diverse exploration strategies. 
However, by approximately episode 15, this variance diminishes substantially, and reward trajectories stabilize, indicating policy convergence.
In both cases, the accumulative rewards demonstrate a consistent upward trajectory with diminishing volatility, confirming systematic improvement in decision quality.
They provide compelling evidence of effective learning.

\subsection{RL Agent Inference Evaluation}
This section presents a comprehensive analysis of \Sysname's performance during the inference phase, wherein pre-trained RL agents actively modulate batch sizes throughout distributed training.
We evaluate the efficacy of our approach across multiple experimental configurations to assess both generalization performance and computational efficiency gains.
To evaluate inference performance, we deploy pre-trained RL agents from Section~\ref{sec:rl_training} across three distinct configurations: VGG11 with SGD, VGG11 with Adam, and ResNet34 with SGD.
For each configuration, we conduct multiple independent experimental runs to mitigate stochastic variability and demonstrate the average inference performance. 
The inference process maintains continuous state-action cycles wherein the RL agent receives multi-dimensional state information, determines optimal batch size adjustments, and implements these adjustments until training convergence.
To ensure comparative validity with the training phase results, each target model (VGG11 and ResNet34) undergoes identical training step sequences as those implemented during RL agent training (Section~\ref{sec:rl_training}).
This enables direct performance comparisons to quantify the generalization capabilities of the trained RL policies.
Our evaluations focus on three key criteria: the final convergence accuracy, which represents the statistical efficiency; the total time-to-convergence, representing the computational efficiency; and the batch size adaptation patterns, which show the optimization dynamics.
We also compare these results with the static batch size baselines established in Section~\ref{sec:baseline}.

\begin{figure*}[t]
    \centering
    \begin{subfigure}[t]{0.32\linewidth}
        \centering
        \includegraphics[scale=0.2]{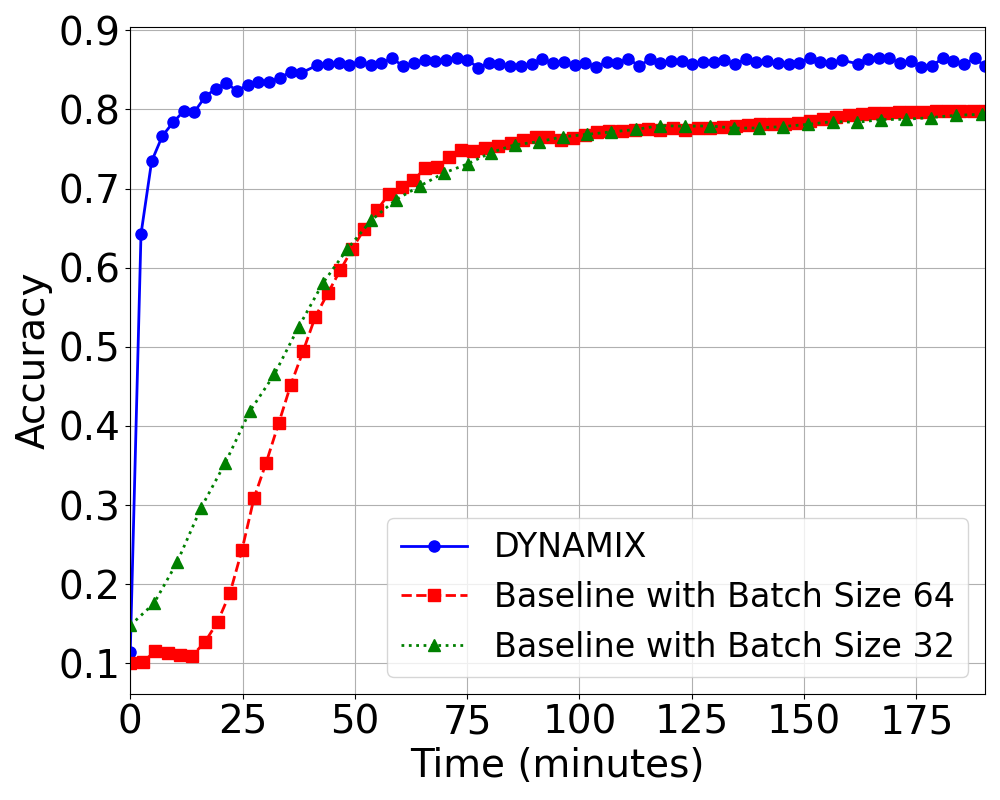}  
        \caption{VGG11 with SGD Optimizer}
        \label{fig:vgg11_inference_accuracy}
    \end{subfigure}
    \hfill
    \begin{subfigure}[t]{0.32\linewidth}
        \centering
        \includegraphics[scale=0.2]{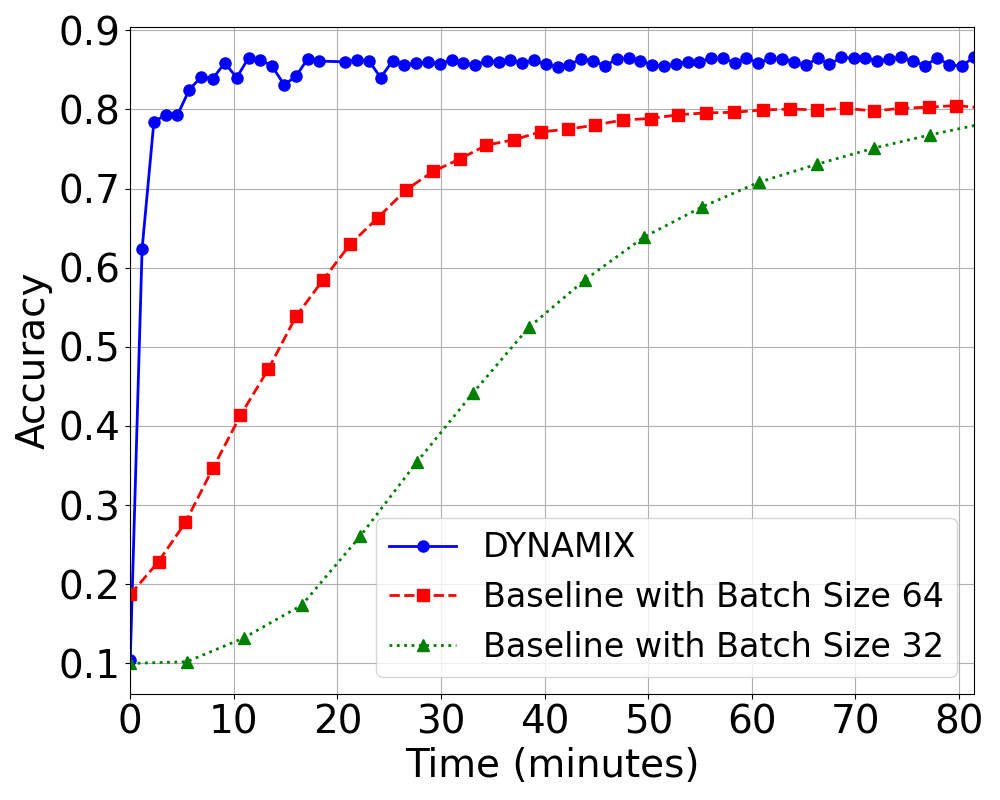} 
        \caption{VGG11 with ADAM Optimizer}
        \label{fig:vgg11_adam_inference_accuracy}
    \end{subfigure}
    \hfill
    \begin{subfigure}[t]{0.32\linewidth}
        \centering
        \includegraphics[scale=0.2]{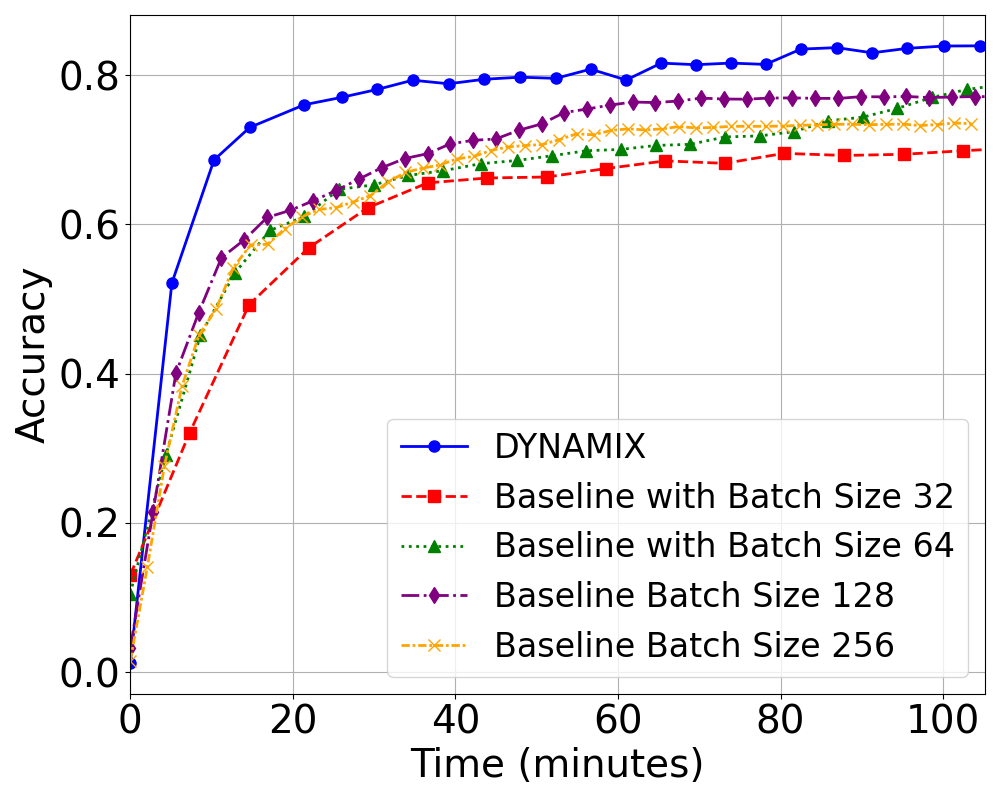}  
        \caption{ResNet34 with SGD Optimizer}
        \label{fig:resnet34_inference_accuracy}
    \end{subfigure}
    \caption{Accuracy trajectories during target model training}
    \label{fig:inference_accuracy}\vspace{-1em}
\end{figure*}

\BfPara{Model Convergence}
Figure~\ref{fig:inference_accuracy} shows the inference accuracy trajectories for VGG11-SGD, VGG11-Adam, and ResNet34-SGD training, respectively.
We note that all the configurations exhibit remarkably consistent performance across runs, with negligible inter-run variance, indicating robust and deterministic decision-making by the RL agent.
So, we randomly select one of the experiment runs to demonstrate the results in the figures.
We can see that the VGG11-SGD configuration demonstrates particularly rapid convergence, achieving 80\% accuracy within approximately 10 minutes and terminal accuracy of 86\% by 30 minutes.
Notably, this convergence rate substantially outperforms the static batch size baseline results shown in Figures~\ref{fig:sgd_32_baseline} and ~\ref{fig:sgd_64_baseline}.
It is clear that VGG11-SGD with \Sysname achieves much higher accuracy (86\%) faster than both the 32-batch and 64-batch static configurations, which require approximately 190 minutes to reach comparable accuracy levels. 
This represents an approximate 6.3x acceleration in convergence time while maintaining equivalent or higher terminal accuracy.

Similar patterns are also shown in the VGG11-Adam configuration~\ref{fig:vgg11_adam_inference_accuracy}.
With \Sysname, it achieves 86\% accuracy within 30 minutes, compared to the 80 minutes required for the static batch size 64 configuration to reach 80\% accuracy \textemdash\xspace a 2.67x improvement with 6\% higher terminal accuracy. 
The ResNet34-SGD configuration (Figure~\ref{fig:resnet34_inference accuracy}) demonstrates comparable acceleration characteristics, with \Sysname converging to 82\% accuracy in 80 minutes, while static configurations require significantly longer training periods to achieve lower terminal accuracy levels.
These results demonstrate that \Sysname effectively addresses the limitations in static batch size approaches.
This adaptive capability enables simultaneous improvements in both convergence rate and terminal accuracy.

\BfPara{Batch Size Adaptation Dynamics}
A distinguishing feature of \Sysname is its capacity to implement non-uniform, temporally-variant batch size strategies.
\begin{figure*}[t]
    \centering
    \begin{subfigure}[t]{0.32\linewidth}
        \centering
        \includegraphics[scale=0.2]{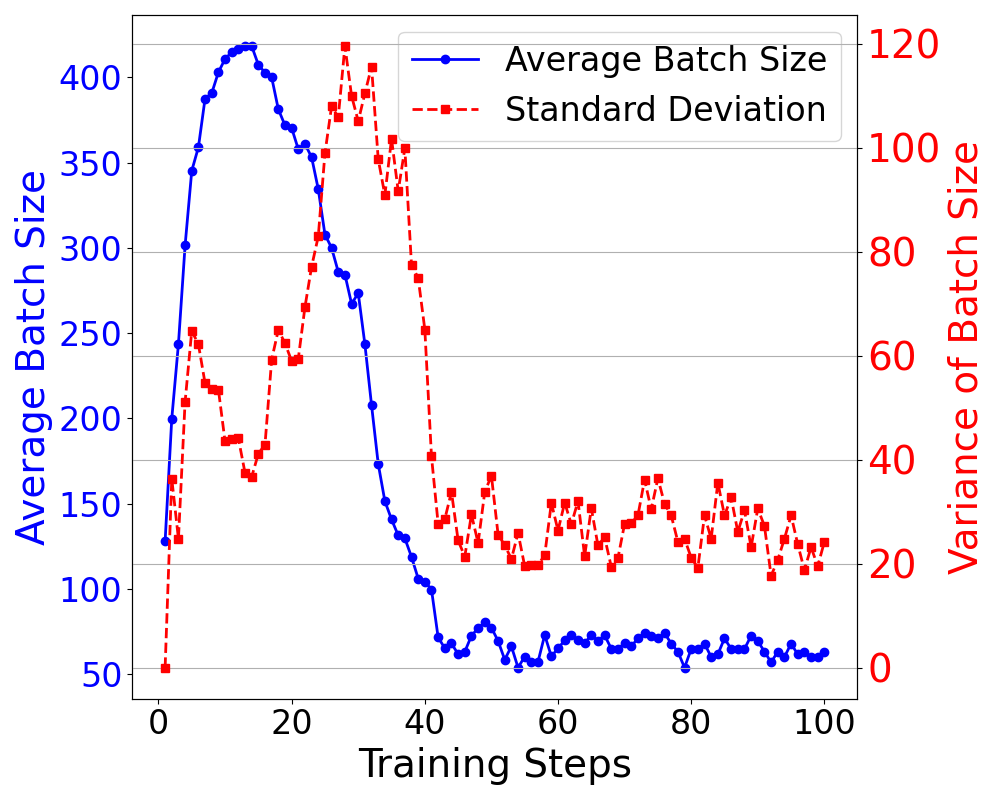}  
        \caption{VGG11 with SGD Optimizer}
        \label{fig:vgg11_sgd_batch}
    \end{subfigure}
    \hfill
    \begin{subfigure}[t]{0.32\linewidth}
        \centering
        \includegraphics[scale=0.2]{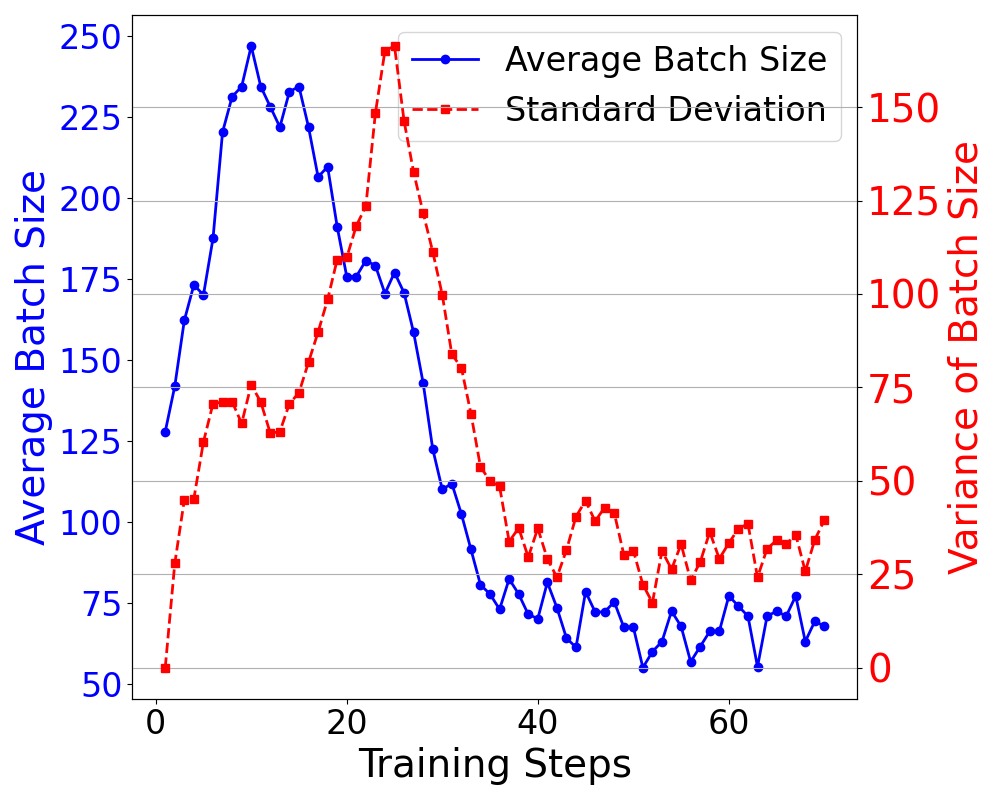} 
        \caption{VGG11 with ADAM Optimizer}
        \label{fig:vgg11_adam_batch}
    \end{subfigure}
    \hfill
    \begin{subfigure}[t]{0.32\linewidth}
        \centering
        \includegraphics[scale=0.2]{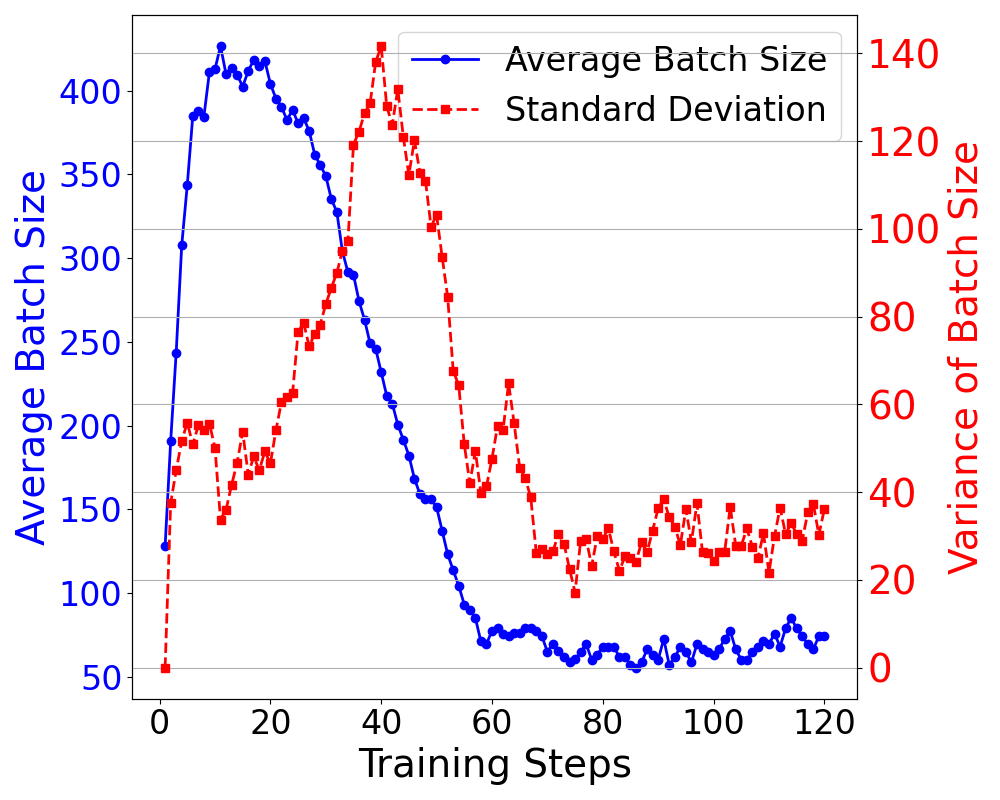}  
        \caption{ResNet34 with SGD Optimizer}
        \label{fig:resnet34_sgd_batch}
    \end{subfigure}
    \caption{Batch size adjustments during target model training}
    \label{fig:batchsize_adjust}\vspace{-2em}
\end{figure*}
Figure~\ref{fig:batchsize_adjust} shows the batch size adjustment trajectories throughout training across all the configurations.
In these figures, we demonstrate both the average batch sizes and their standard deviations to visualize the characteristics of the batch size adjustment.
We can see that across all the configurations, the RL agent initially selects relatively large batch sizes (e.g., $\thicksim$400 for VGG11-SGD and ResNet34-SGD and $\thicksim$250 for VGG11-ADAM) to accelerate early training through increased parallelism, subsequently transitions to medium batch sizes during mid-training to balance hardware utilization with gradient quality, and ultimately reduces batch sizes during final convergence phases to enhance fine-grained optimization.
Similar trends can also be observed for the standard deviation changes.
These trends align with the recent theoretical work suggesting that gradient noise characteristics vary systematically throughout training, requiring correspondingly adaptive batch size configurations to optimize convergence properties. 
For example, Smith \etal~\cite{smith2017don} demonstrate that gradient noise scale evolves predictably during training, establishing a theoretical foundation for dynamic batch sizing approaches. 
The observed three-phase pattern (large$\rightarrow$medium$\rightarrow$small) represents an empirically derived optimization strategy that effectively navigates the time-varying statistical-computational trade-off landscape inherent in distributed training.

\BfPara{Comparative Performance}
When comparing with the training results from Section~\ref{sec:rl_training}, inference performance exhibits several notable characteristics. 
The comparison of the accuracy trajectories with the baseline results are also shown in Figure~\ref{fig:inference_accuracy}.
We can see that \Sysname achieves higher accuracy and converges faster than using the fixed batch sizes across all training configurations.
Further, the accelerated convergence observed during inference surpasses even the later-episode training performance, suggesting effective generalization of the learned policy to novel training trajectories.
Furthermore, the inter-run consistency of inference results substantially exceeds that observed during training, indicating policy stabilization and robustness.
The batch size adjustment pattern also confirms the successful transfer of optimization strategy from training to inference contexts.
These results empirically validate the efficacy of our reinforcement learning approach in addressing the fundamental limitations of static batch size allocation in heterogeneous distributed training environments.

\subsection{Scalability of \Sysname}
Scalability is a critical property for any distributed learning system, as modern deep learning workloads increasingly require training across dozens or hundreds of computational nodes to achieve acceptable training times.
However, as the number of nodes increases, gradient synchronization often becomes more complex, floating-point rounding errors accumulate during all-reduce operations, and communication overhead grows substantially.
To validate \Sysname's practical applicability in production environments, we conduct scalability experiments using VGG16 on CIFAR-10 with SGD across 8, 16, and 32 nodes on the OSC cluster. 
To establish meaningful comparisons, we identify the optimal static batch size configuration for each cluster scale through systematic evaluation, and then compare \Sysname's dynamic optimization against these baselines.
The results are shown in Table~\ref{tab:scalability}.
\begin{table}[h]
\caption{Scalability of \Sysname}\label{tab:scalability}
\resizebox{\linewidth}{!}{
\renewcommand{\arraystretch}{1.2}
\begin{tabular}{|c|ccc|cc|}
\hline
\multirow{2}{*}{\textbf{Cluster Size}} & \multicolumn{3}{c|}{\textbf{Static Batch Size}} & \multicolumn{2}{c|}{\textbf{\Sysname}} \\ \cline{2-6} 
 & Batch size & Accuracy & ConvTime (sec) & Accuracy & ConvTime (sec) \\ \hline
8 Nodes & \multicolumn{1}{c|}{128} & \multicolumn{1}{c|}{85.3\%} & 853 & \multicolumn{1}{c|}{\textbf{91.3\%}} & \textbf{652} (\textcolor{red}{$\downarrow30.1$\%}) \\ \hline
16 Nodes & \multicolumn{1}{c|}{128} & \multicolumn{1}{c|}{83.4\%} & 543 & \multicolumn{1}{c|}{\textbf{91.5\%}} & \textbf{479} (\textcolor{red}{$\downarrow 13.7$\%}) \\ \hline
32 Nodes & \multicolumn{1}{c|}{64} & \multicolumn{1}{c|}{81.3\%} & 734 & \multicolumn{1}{c|}{\textbf{92.6\%}} & \textbf{421} (\textcolor{red}{$\downarrow 42.6$\%}) \\ \hline
\end{tabular}
}
\end{table}
From the table, we can see that with static batch sizes, there exists a clear accuracy degradation as cluster size increases, decreasing from 85.3\% at 8 nodes to 81.3\% at 32 nodes.
Furthermore, the optimal static batch size changes across scales, indicating that static batch sizes cannot maintain consistent optimization strategies. 
In contrast, \Sysname improves both accuracy and convergence time as cluster size increases.
The final model accuracy increases consistently across scales, suggesting that \Sysname can better exploit the increased computational resources while mitigating the coordination challenges that degrade static approaches.
Similar for convergence time, \Sysname achieves significant reductions across scales.
These results demonstrate that \Sysname addresses one of the most persistent challenges in distributed learning systems: the accuracy-scalability trade-off that typically forces practitioners to choose between training speed and model quality as cluster size increases.

\subsection{Policy Transferability Across Model Architectures}
\begin{figure}[h]
    \centering
    \includegraphics[scale=0.7]{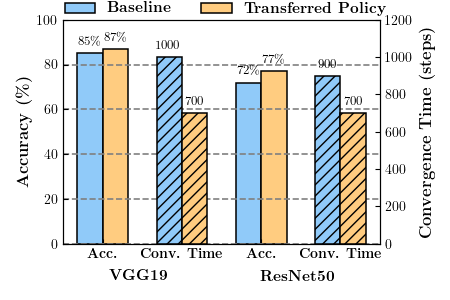}\vspace{-1em}
    \caption{Performance of transferred policies}
    \label{fig:transferrability}
\end{figure}

In practice, engineers may work with multiple model variants within the same architectural family, such as VGG16 and VGG19.
Training separate reinforcement learning policies for each model variant would impose significant computational overhead and delay deployment timelines.
The ability to transfer learned batch size optimization policies across related architectures would be a critical practical advantage for real-world adoption of \Sysname.
If policies learned on one model can effectively generalize to deeper or wider variants within the same family, this would dramatically reduce the training overhead.
Therefore, we conduct two experiments to evaluate policy generalizability across model families with different computational characteristics and dataset complexities.
For the first experiment, the source policy is trained using VGG16 on the CIFAR-10 dataset with SGD optimization.
Then, the trained policy is directly applied to guide VGG19 training under identical system configurations, hardware resources, and optimization settings.
Similarly, we also train a dynamic scheduling policy on ResNet34 using the CIFAR-100 dataset and apply the learned policy to ResNet50 training.
Both experiments are conducted on the OSC cluster.
VGG16 and VGG19 training utilize 16 nodes, while ResNet34 and ResNet50 training utilize 32 nodes.
We compare the transferred policy's performance against the carefully tuned baselines \textendash\xspace the optimal static batch size configurations for each target architecture, and present the results in Figure~\ref{fig:transferrability}.
From the figure, we can see that the learned policies for both VGG16 and ResNet34 are successfully generalized to VGG19 and ResNet50.
They not only improve the final accuracy of the models, but also significantly reduce the convergence time compared with the baseline approaches.
The consistent improvements indicate that \Sysname successfully captures and leverages architectural similarities within model families.
This capability dramatically reduces the deployment overhead for \Sysname in production environments where multiple model variants are commonly evaluated.
While our experiments focus on intra-family transfers, the results motivate future investigation into cross-family transfers and the broader applicability of learned batch size optimization strategies.

\subsection{Integration with BytePS}
To validate \Sysname's framework-agnostic capabilities and resilience to hardware heterogeneity, we conduct experiments to demonstrate its adaptability across different distributed training architectures.
To that end, we utilize BytePS~\cite{jiang2020unified}, which is a high-performance distributed deep learning framework that employs a parameter server architecture for gradient synchronization across worker nodes.
The experimental cluster consists of 8 heterogeneous GPU worker nodes.
An additional CPU-only node hosts the \Sysname scheduler, communicating with all GPU nodes via standard TCP networking.
Then, we train a VGG11 model on the CIFAR-10 dataset with SGD optimization.
For the baseline, we train the model with a static batch size of 64.
It takes about 20,000 seconds for the training to converge at 71.4\% final accuracy.
With \Sysname, the training converges after about 16,000 seconds to reach  80\% final accuracy, which is 8.6\% improvement on accuracy and 20\% reduction in time.
The seamless integration with BytePS demonstrates that \Sysname's optimization principles generalize effectively across different distributed training paradigms. 
Furthermore, \Sysname successfully adapts to a heterogeneous environment, which represents realistic production deployment conditions.
These results show that \Sysname is a hardware-agnostic, framework-independent solution capable of delivering consistent performance improvements.

\subsection{Overhead Analysis}
Our empirical analysis demonstrates that the additional overhead introduced by eBPF and gRPC remains negligible compared to total training step time, even under varying load conditions.
We conduct systematic measurements of the time required for the entire system to compute and propagate batch size adjustment decisions. 
Results consistently show that the decision-making overhead represents less than 0.1\% of typical iteration time across all tested configurations. 
The lightweight nature of eBPF kernel programs and the efficient gRPC communication protocol ensure that \Sysname can be safely integrated into distributed training pipelines without creating bottlenecks or compromising performance.
\section{Conclusion}
\label{sec:conclusion}
\Sysname proposes an RL-based framework for adaptive batch size optimization in DML systems. 
Traditional static or heuristic-based batch sizing approaches often fail to adapt to the heterogeneous and dynamic nature of modern computing environments. 
By formulating batch size tuning as a sequential decision-making problem, \Sysname leverages a centralized PPO agent to adjust per-node batch sizes based on a multi-dimensional state representation, including system-level metrics, network-level statistics, and training efficiency indicators. 

Extensive experiments demonstrate that \Sysname achieves faster convergence, better final accuracy, and reduced variance compared to fixed-batch baselines.

\bibliographystyle{IEEEtranS}
\bibliography{ref}

\end{document}